\documentclass{article}

\usepackage{amsmath}
\usepackage[hidelinks]{hyperref}
\usepackage{graphicx}
\PassOptionsToPackage{numbers}{natbib}

\usepackage[preprint]{neurips_2024}

\usepackage[utf8]{inputenc} % allow utf-8 input
\usepackage[T1]{fontenc}    % use 8-bit T1 fonts
\usepackage{hyperref}       % hyperlinks
\usepackage{url}            % simple URL typesetting
\usepackage{booktabs}       % professional-quality tables
\usepackage{amsfonts}       % blackboard math symbols
\usepackage{nicefrac}       % compact symbols for 1/2, etc.
\usepackage{microtype}      % microtypography
\usepackage{xcolor}         % colors

\title{ICLERB: In-Context Learning Embedding and Reranker Benchmark}

\author{%
  Marie Al Ghossein\thanks{Equal contributions. Correspondence should be addressed to \texttt{emile@crossingminds.com}.} \\
  Crossing Minds, Inc \\
  \And
  Emile Contal\footnotemark[1] \\
  Crossing Minds, Inc \\
  \And
  Alexandre Robicquet \\
  Crossing Minds, Inc \\
}

\begin{document}

\maketitle

\begin{abstract}
In-Context Learning (ICL) enables Large Language Models (LLMs) to perform new tasks by conditioning on prompts with relevant information. Retrieval-Augmented Generation (RAG) enhances ICL by incorporating retrieved documents into the LLM's context at query time. However, traditional retrieval methods focus on semantic relevance, treating retrieval as a search problem. In this paper, we propose reframing retrieval for ICL as a recommendation problem, aiming to select documents that maximize utility in ICL tasks. We introduce the In-Context Learning Embedding and Reranker Benchmark (ICLERB), a novel evaluation framework that compares retrievers based on their ability to enhance LLM accuracy in ICL settings. Additionally, we propose a novel Reinforcement Learning-to-Rank from AI Feedback (RLRAIF) algorithm, designed to fine-tune retrieval models using minimal feedback from the LLM. Our experimental results reveal notable differences between ICLERB and existing benchmarks, and demonstrate that small models fine-tuned with our RLRAIF algorithm outperform large state-of-the-art retrieval models. These findings highlight the limitations of existing evaluation methods and the need for specialized benchmarks and training strategies adapted to ICL.
\end{abstract}

\section{Introduction}
\label{sec:introduction}

\textbf{Large Language Models} (LLMs) have demonstrated remarkable capabilities across a wide range of natural language processing tasks, including question answering, text completion, and summarization~\cite{brown2020language}. Their ability to generate coherent and contextually appropriate text has made them invaluable tools in research and industry applications. However, leveraging LLMs for specific domains or tasks often requires injecting domain-specific knowledge and performing adaptations to meet the demands of particular use cases.

Knowledge injection into LLMs is traditionally achieved through fine-tuning, where the model's parameters are updated using task-specific data~\cite{howard2018universal}. While effective, fine-tuning presents several challenges, such as the risk of catastrophic forgetting~\cite{Kirkpatrick_2017}, where the model loses previously learned general knowledge in favor of new task-specific information. Moreover, fine-tuning large models usually demands large labeled datasets and results in substantial operational costs from high computational requirements.

\textbf{In-Context Learning} (ICL) offers a promising alternative by enabling LLMs to perform new tasks through conditioning on prompts that include relevant demonstrations or documents without updating the model's parameters~\cite{brown2020language, minEMNLP2022}. ICL leverages the LLM's ability to adapt to new tasks by incorporating contextual information directly into the input prompt. This approach mitigates the risks associated with fine-tuning and provides flexibility in rapidly adapting models to diverse tasks and domains.

\pagebreak  % move if needed
\textbf{Retrieval-Augmented Generation} (RAG) further enhances ICL by incorporating retrieved documents into the LLM's context, supplying it with up-to-date or domain-specific knowledge that it may lack~\cite{lewis2021retrieval, izacard2022atlas}. RAG systems dynamically retrieve relevant information based on the query and include it in the prompt, potentially improving performance by providing external knowledge to ground the LLM's responses.
In \cite{ovadia-etal-2024-fine} the authors show that retrieval consistently outperforms fine-tuning in a diverse set of knowledge injection tasks.
Central to the efficacy of RAG in ICL is the retrieval model, which selects documents that can effectively enhance the LLM's performance on the given query.

Existing retrieval methods frame the retrieval task as a \textit{search problem} when deriving evaluation frameworks or training datasets and objectives. These methods focus on retrieving documents that match queries based on relevance. "Relevance" in this context often indicates semantic proximity without necessarily considering the document's utility in improving the LLM's task performance. 

The wide adoption of this formulation is particularly evident when looking at the datasets used for evaluating and fine-tuning retrieval models for the two tasks of retrieval and ranking. These datasets are typically designed for search tasks, and when used for evaluation, they allow one to measure how effectively models retrieve or rank documents based on semantic relevance to search queries. Notable benchmarks include the BEIR~\cite{thakur2021beir} benchmark and the Massive Text Embedding Benchmark (MTEB)~\cite{muennighoffEACL2023}. Examples of retrieval datasets include DBpedia Entity~\cite{Balog2013DBpedia}, containing entity-oriented search queries, MS MARCO~\cite{bajaj2018msmarco}, comprising search web queries from Bing and passages from relevant documents, and Quora~\cite{thakur2021beir}, consisting of pairs of questions labeled as semantically equivalent or not. Examples of ranking datasets include StackOverflowDupQuestions~\cite{liu2018linkso}, a dataset for learning to rank similar questions on Stack Overflow, and MindSmall~\cite{wu2020mind}, a news recommendation dataset. 

Moreover, RAG systems often employ off-the-shelf pre-trained text embedding models and then use cosine similarity for retrieval~\cite{reimers-2019-sentence-bert}, heavily relying on semantic similarity. Fine-tuning for retrieval tasks is also typically performed on search datasets; for example, ColBERT~\cite{khattabSIGIR2020} is fine-tuned on MS MARCO to improve retrieval performance.

However, we argue that the RAG retrieval problem is a \textit{recommendation problem}, instead of being a search problem. The primary objective of ICL is not merely to find documents that are semantically similar to the query, but to retrieve documents ("items" in recommender systems) that provide maximal utility given the query ("session" in recommender systems).

Evaluating retrievers in this context should therefore be based on downstream metrics that directly measure how much the retrieved documents increase the accuracy of the enriched LLM. Traditional evaluation metrics that focus on relevance or semantic similarity may not capture this aspect, requiring new benchmarks and evaluation methodologies. In light of this, we introduce the In-Context Learning Embedding and Reranker Benchmark (ICLERB) as a new evaluation framework that implements these ideas and evaluates retrievers for the task of ICL. 

It is important to note that fine-tuning retrieval models to optimize for utility in ICL presents significant challenges. Generating a retrieval dataset that reflects the utility of documents in improving LLM performance is computationally expensive. Evaluating all possible query-document pairs is infeasible due to the quadratic number of pairs, and naive approaches that rely on subsampling are insufficient. Typically, using weighted sampling based on an unsupervised scorer is limited by the fact that the scorer might not align with the objectives of ICL. 

To address these challenges, we introduce a \textbf{Reinforcement Learning-to-Rank from AI Feedback} (RLRAIF) algorithm. This novel approach fine-tunes retrieval models using feedback from the LLM itself while optimizing a novel dual exploration/exploitation trade-off. RLRAIF enables the retriever to learn which documents are most useful in enhancing LLM performance, directly optimizing for the retrieval problem as a recommendation task and reaching superior performance with minimal budget of LLM queries.

\begin{figure}[htbp]
    \centering
    \includegraphics[width=1\textwidth]{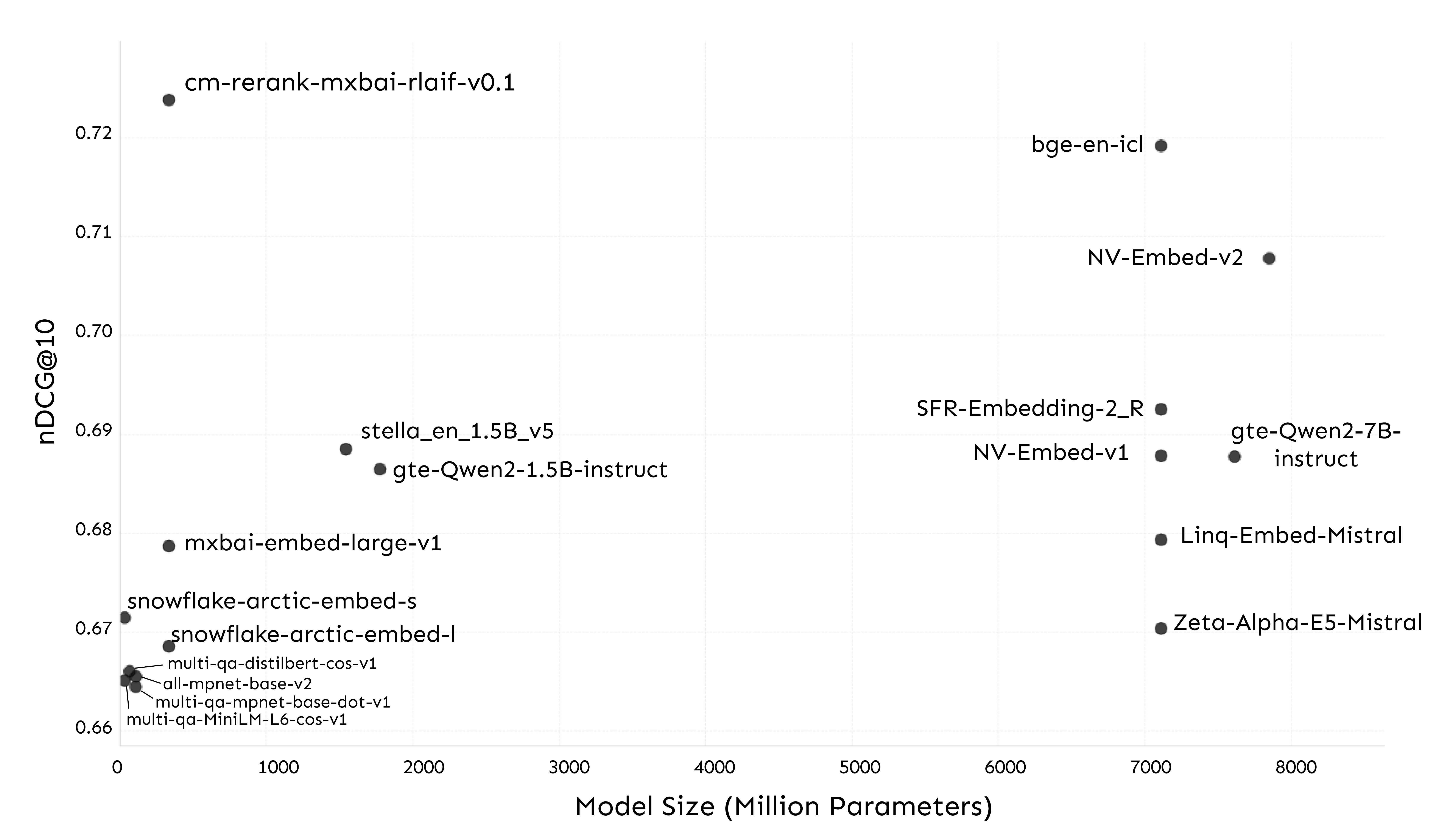}
    \caption{Retrieval performance for the task of ICL, as measured by nDCG@10 in the In-Context Learning Embedding and Reranker Benchmark (ICLERB), vs. model size in millions of parameters. \texttt{cm-rerank-mxbai-rlaif-v0.1} (ours) is fine-tuned by RLRAIF and outperforms other approaches while being considerably smaller than most models.}
    \label{fig:ndcg_vs_size}
\end{figure}

In this paper, we will address the following two research questions:
\begin{itemize}
    \item \textbf{RQ1.} How does the performance of embedding models for ICL compare to traditional benchmarks like MTEB~\cite{muennighoffEACL2023}?
    \item \textbf{RQ2.} Can we fine-tune a retrieval model for ICL without requiring a large number of LLM queries?
\end{itemize}

\pagebreak  % might need to be moved
Our contributions are threefold:

\begin{enumerate}
    \item \textbf{Novel Evaluation Methodology}: We introduce a methodology to evaluate retrievers for ICL based on how much the retrieved documents enhance LLM performance. This approach provides a more relevant assessment compared to traditional benchmarks focused on semantic similarity.
    \item \textbf{ICLERB Benchmark}: We present the In-Context Learning Embedding and Reranker Benchmark (ICLERB), which ranks the performance of existing embedding models and rerankers across multiple datasets and LLMs using our proposed evaluation methodology.
    \item \textbf{RLRAIF Approach}: We propose a new approach based on Reinforcement Learning-to-Rank from AI Feedback (RLRAIF) that fine-tunes retrieval models directly for the task of ICL using RL. We demonstrate the merits of RLRAIF in the context of the first release of ICLERB. Our model, \texttt{cm-rerank-mxbai-rlaif-v0.1}, based on the \texttt{mxbai-embed-large-v1} pretrained embeddings, shows significant improvements in retrieval performance for ICL tasks while being of notably smaller size compared to most methods (Figure~\ref{fig:ndcg_vs_size}).
\end{enumerate}

The rest of the paper is organized as follows: Section~\ref{sec:related_work} presents a comprehensive review of related work, followed by a detailed description of ICLERB in Section~\ref{sec:iclerb}. We then discuss the ICLERB results in Section~\ref{sec:iclerb_results} and introduce our new RLRAIF approach in Section~\ref{sec:rlaif}. 

The ICLERB leaderboard is available on HuggingFace\footnote{\url{https://huggingface.co/spaces/crossingminds/iclerb}} and will be regularly updated with future comparisons and extensions of this work.
We aim to release all the code required to reproduce ICLERB on GitHub\footnote{\url{https://github.com/Crossing-Minds}} in the near future.
Up-to-date related resources can be found on the ICLERB web page\footnote{\url{https://www.crossingminds.com/iclerb}}.

\section{Related Work}
\label{sec:related_work}

In this section, we review existing research pertinent to improving retrieval methods for In-Context Learning (ICL) with Large Language Models (LLMs). We highlight the limitations of unsupervised retrieval approaches based on Semantic Text Similarity (STS) and discuss supervised methods that aim to optimize retrieval for ICL tasks. We also underscore the absence of benchmarks specifically designed to evaluate retrievers in the context of ICL, motivating the development of our proposed benchmark.

\subsection{In-Context Learning and Retrieval-Augmented Generation}

\textbf{In-Context Learning (ICL)} refers to the ability of LLMs to perform tasks by conditioning on prompts that include guidance, demonstration, documents, or context, without updating the model parameters~\cite{QingxiuARXIV2023}. LLMs exhibit meta-learning capabilities, effectively employing inductive reasoning on the information from the prompt, enabling zero-shot or few-shot learning~\cite{minEMNLP2022}. This capacity has been linked to statistical inference, where the LLM acts as an in-context optimizer for problems like least squares, ridge regression, or gradient descent~\cite{BaiNEURIPS2023}.

\textbf{Retrieval-Augmented Generation (RAG)} enhances ICL by incorporating retrieved documents into the LLM's context~\cite{RamTACL2023}. Unlike static few-shot learning or prompt engineering, RAG dynamically retrieves relevant information based on the query and includes it in the prompt, potentially improving performance by providing up-to-date or domain-specific knowledge that the LLM may lack~\cite{IzacardEACL2021}. This approach leverages external knowledge bases to ground the LLM's responses, enhancing its ability to generate accurate and contextually appropriate outputs. A particular application of RAG is dynamic few-shot learning, where documents are demonstrations of ground truth query-response pairs.
Existing RAG systems typically utilize retrieval methods designed for similarity search tasks, focusing on retrieving documents that are semantically similar to the query~\cite{KarpukhinEMNLP2020}.

\subsection{Limitations of Unsupervised Retrieval Methods Based on Semantic Similarity}

\subsubsection{Text Embeddings and Cosine Similarity}
\label{subsec:related_work_text_embs_cosine_sim}

Text embeddings are fundamental to many Natural Language Processing (NLP) tasks, serving as continuous vector representations that capture semantic and syntactic properties of text~\cite{mikolov2013distributed}. Pre-trained language models like BERT~\cite{devlinNAACL2019} and T5~\cite{raffelJMLR2020} have been widely used to generate embeddings for tasks such as classification, retrieval, and question answering.

In RAG systems, a common approach is to use off-the-shelf pre-trained embeddings and cosine similarity to model document relevance~\cite{KarpukhinEMNLP2020, IzacardEACL2021}. Given a query $q$ and a set of documents $D$, relevance is computed as:

\begin{equation}
\operatorname{rel}(q, d) = \cos(\mathbf{e}_q, \mathbf{e}_d) = \frac{\mathbf{e}_q \cdot \mathbf{e}_d}{\|\mathbf{e}_q\| \|\mathbf{e}_d\|}, \quad \forall d \in D,
\end{equation}

where $\mathbf{e}_q$ and $\mathbf{e}_d$ are the embeddings of the query and document, respectively.

Using embedding distances for ICL retrieval is widely studied in the context of few-shot learning, where documents are demonstrations of desired responses~\cite{rubinNACL2022, wangArXiv2022, liACL2023, yeICML2023, wangEACL2024}. For instance, \verb|KATE|~\cite{liuDEELIO2022} retrieves a set of $k$ documents using $k$-Nearest Neighbors (\verb|kNN|) search with cosine similarity, leveraging vector search algorithms for efficient retrieval in large-scale settings.

\subsubsection{Pre-trained Embeddings for Semantic Similarity}

Approaches like Sentence-BERT (SBERT)~\cite{reimers-2019-sentence-bert} and SimCSE~\cite{gaoEMNLP2021} learn embeddings by fine-tuning pre-trained models on supervised datasets that focus on Semantic Text Similarity (STS). SBERT modifies the BERT architecture to produce high-quality sentence embeddings by incorporating a pooling operation over token embeddings and fine-tuning on datasets like the Stanford Natural Language Inference (SNLI)~\cite{bowmanEMNLP2015} and the Multi-Genre Natural Language Inference (MultiNLI)~\cite{williamsNAACL2018}, where the ground truth involves recognizing entailment, contradiction, or neutrality between sentence pairs.

The training process involves using labeled pairs of semantically similar and dissimilar sentences to enforce that embeddings of similar sentences are close in the vector space, while those of dissimilar sentences are distant. The contrastive learning objective typically minimizes a loss function such as:

\begin{equation}
\mathcal{L}_{\text{contrastive}} = - \sum_{i,j} y_{ij} \log \sigma(\cos(\mathbf{e}_i, \mathbf{e}_j)) + (1 - y_{ij}) \log (1 - \sigma(\cos(\mathbf{e}_i, \mathbf{e}_j))),
\end{equation}

where $y_{ij} \in \{0,1\}$ indicates whether sentences $i$ and $j$ are semantically similar, and $\sigma(\cdot)$ is the sigmoid function.

Similarly, SimCSE employs both supervised and unsupervised contrastive learning to produce embeddings. In the unsupervised setting, positive pairs are created by applying dropout noise to the same input sentence, encouraging the model to produce similar embeddings for different augmentations of the same sentence. The loss function used is the normalized temperature-scaled cross-entropy loss~\cite{chenICML2020}:

\begin{equation}
\mathcal{L}_{\text{NT-Xent}} = - \sum_{i} \log \frac{ e^{\cos(\mathbf{e}_i, \mathbf{e}_i^+) / \tau} }{ \sum_{j} e^{\cos(\mathbf{e}_i, \mathbf{e}_j^+) / \tau} },
\end{equation}

where $\mathbf{e}_i^+$ is the positive pair for $\mathbf{e}_i$, and $\tau$ is a temperature parameter.

\subsubsection{Pre-training with Weak Supervision on Semantic Similarity}

To overcome the necessity for large supervised data sets, several approaches involve pre-training models on large-scale weakly supervised or unlabeled datasets using contrastive learning. Notable examples include E5~\cite{wangArXiv2022}, GTE~\cite{liCoRR2023}, OpenAI's embeddings~\cite{neelakantanArXiv2022}, BGE~\cite{xiaoSIGIR2024}, and NV-Embed~\cite{lee2024nv}.

These methods often employ strategies like the Inverse Cloze Task (ICT)~\cite{changICLR2020}, where a sentence is extracted from a passage to serve as a query, and the remaining text constitutes the positive context. The goal is to train the model to retrieve the original context given the query, emphasizing local semantic coherence.

Other techniques include random cropping~\cite{izacardTMLR2023} and using neighboring text spans~\cite{neelakantanArXiv2022} to create positive and negative pairs for contrastive learning. For example, given a query $q_i$, a positive document $d_i^+$, and negative documents $\{d_i^-\}$, the model maximizes:

\begin{equation}
p(d_i^+ | q_i) = \frac{ e^{\cos(\mathbf{e}_{q_i}, \mathbf{e}_{d_i^+}) / \tau} }{ e^{\cos(\mathbf{e}_{q_i}, \mathbf{e}_{d_i^+}) / \tau} + \sum_{d \in \{d_i^-\}} e^{\cos(\mathbf{e}_{q_i}, \mathbf{e}_{d}) / \tau} }.
\end{equation}

However, these pre-training methods still focus on semantic similarity or contextual relevance, potentially neglecting aspects that are critical for ICL utility. The positive examples are constructed based on their semantic relation to the query rather than their effectiveness in improving LLM performance in specific tasks.

Moreover, the techniques for mining hard negatives often rely on the assumption that negative examples are semantically dissimilar, which may not hold true in the context of ICL~\cite{ni2021large}. In ICL, hard negatives could be examples that are semantically similar but do not contribute to performance improvement, highlighting the need for utility-based negative mining strategies.

\subsubsection{Rerankers Trained on Binary Query Relevance}
\label{subsec:related_work_rankers}

Rerankers, or simply rankers, are neural ranking models that encode a query and a document together to directly predict the relevance. Instead of modeling the relevance as a dot product, they typically involve non-linearity, making it harder to index using tools like FAISS \cite{douze2024faiss} or vector databases. The name reranker comes from a two-stage process where candidates are first selected using indexed retrieval based on embeddings, and then candidates are ranked using a relevance model.

Models like ColBERT~\cite{khattabSIGIR2020} introduce late interaction mechanisms that allow for fine-grained matching between queries and documents at the token level. Rerankers such as Cohere's reranker~\cite{cohereRerank} and Voyage AI's \texttt{rerank}~\cite{voyageaiRerank} utilize cross-encoder architectures that jointly encode the query and document, capturing complex interactions missed in bi-encoder setups.

These rerankers are typically trained on datasets where relevance is judged based on semantic or topical similarity, focusing on predicting relevance judgments rather than enhancing ICL utility. Their optimization objectives do not consider the utility of documents in improving LLM performance in specific tasks. As a result, they may not prioritize documents that are most beneficial when used as in-context examples for LLMs.

\subsubsection{Limitations of Semantic Similarity for RAG ICL}

While these methods are effective for a wide range of generic tasks emphasizing binary semantic similarity, they may not select the most useful documents for ICL. Studies have shown that the quality of in-context examples significantly affects LLM performance~\cite{liuDEELIO2022, minEMNLP2022}. Semantically similar documents may not provide the necessary information for optimizing task performance in ICL.

These limitations arise because:

\begin{itemize}
    \item \textbf{Misalignment with ICL Objectives}: Semantic similarity does not necessarily correlate with the usefulness of an in-context example for improving LLM performance~\cite{liuDEELIO2022}. The most semantically similar documents may be of poor quality, or simply not informative.
    \item \textbf{Binary Relevance Labels}: Using binary labels in contrastive learning fails to capture the continuous and task-specific nature of utility in ICL~\cite{RamTACL2023}.
    \item \textbf{Lack of Specificity}: General purpose embeddings and rerankers are trained to perform well on a wide range of tasks such as classification, search or paraphrasing. Specific retrieval models are needed to improve the performances of LLMs using ICL.
\end{itemize}

\subsection{Retrieval Methods Supervised for Few-Shot ICL}
\label{subsec_related_work:retrieval_models}

To address the limitations of unsupervised methods, recent approaches involve supervising retrieval models directly on the downstream ICL task.
This literature typically focuses on the particular case of RAG where documents are demonstrations, also known as dynamic few-shot learning.
Conveniently, the same dataset is used both to measure the performances of the LLM as well as to fine-tune a retrieval model.
Examples include Efficient Prompt Retriever (EPR)~\cite{rubinNACL2022}, Unified Demonstration Retriever (UDR)~\cite{liACL2023}, and LLM-Retriever (LLM-R)~\cite{wangEACL2024}.
These methods train embedding models or rankers to select in-context examples that maximize LLM performance on specific tasks.

\subsubsection{Retrieval Models Supervised on Downstream Tasks}

EPR trains a small language model to predict which demonstrations (or "few-shot examples") are most helpful. It constructs a training dataset in two stages, by first querying the top $L$ examples ranked by an unsupervised retrieval model (e.g. BM25 or SBERT), and then scoring the examples using the small LM to get true utility scores (log probabilities of correct responses). To train the language model using the contrastive loss, EPR converts them to binary labels by forming $k$ positive examples, the ones maximizing the score, and $k$ negative examples, minimizing the scores from the top $L$ candidates:

\begin{equation}
\mathcal{L}_{\text{EPR}} = - \sum_{i} \log \frac{ e^{\cos(\mathbf{e}_q, \mathbf{e}_{d_i^+})} }{ e^{\cos(\mathbf{e}_q, \mathbf{e}_{d_i^+})} + e^{\cos(\mathbf{e}_q, \mathbf{e}_{d_i^-})} },
\end{equation}

where $L$ and $k$ are hyper-parameters, and $d_i^+$ and $d_i^-$ are positive and negative examples, respectively.

UDR pre-trains a single embedding model for multiple retrieval tasks, using an inner product similarity and a ranking loss inspired by LambdaRank~\cite{NIPS2006_af44c4c5}, removing the need to binarize the utility scores. It extends the two-stage approach into several iterations. After a first round where candidates are initially selected using an unsupervised model, UDR employs the trained model itself to mine hard negative examples and build a new dataset used to train the next iteration of the model. In practice the authors limit the number of iterations to 3, since each round requires to query the language model many times to get enough utility scores to build the training dataset.

\begin{equation}
\mathcal{L}_{\text{UDR}} = \sum_{i} \sum_{j,k} \max(0, \Delta_{jk}) \cdot \log\left(1 + e^{ \cos(\mathbf{e}_q, \mathbf{e}_{d_k}) - \cos(\mathbf{e}_q, \mathbf{e}_{d_j}) }\right),
\end{equation}

where $\Delta_{jk} = \frac{1}{r(d_j)} - \frac{1}{r(d_k)}$, $r(d)$ is the rank of document $d$ based on its utility, and $\mathbf{e}_q$, $\mathbf{e}_{d_j}$, and $\mathbf{e}_{d_k}$ are embeddings of the query and documents.

LLM-R follows EPR using binary contrastive loss, but trains a large language model retriever. Like UDR, it employs up to 3 iterations to mine hard negative examples.

The authors demonstrate that supervising the retriever on the downstream task can lead to improved ICL performance. However, they rely on iteratively mining datasets with hard-negatives, constructed from the LLM's output probabilities. Given that large language models typically have billions of parameters, they require large training datasets, which can be computationally intensive to generate.

\subsubsection{Challenges in Evaluating Retrievers Separately}

In these supervised methods, measuring the effectiveness of the retriever only via the performances of the LLM enriched with few shot learning is intertwined with other factors such as prompt optimization techniques, demonstration re-ordering, and diversity of the demonstrations~\cite{lu2022fantastically, contalArxiv2024}. Evaluating the retriever in isolation becomes difficult because improvements in LLM performance may be attributed to aspects beyond the retriever itself.

Without standardized benchmarks that evaluate retrievers based solely on their contribution to ICL performance, it is challenging to compare different retrieval methods and drive progress in this area. There is a pressing need for benchmarks that evaluate retrievers independently, focusing on their ability to select in-context examples that enhance LLM performance.

% TODO - add transition to this next section?
\subsection{Fine-Tuning and DPO}
\label{subsec:fine_tuning_dpo}

\paragraph{Direct Preference Optimization (DPO)}

Originally introduced as an alternative to Reinforcement Learning from Human Feedback (RLHF), DPO is a method for fine-tuning language models directly from human preferences without explicit reward modeling or reinforcement learning~\cite{rafailovNEURIPS2023}. It operates by increasing the relative log probabilities of preferred responses over dispreferred ones, effectively optimizing the model to produce outputs that align with human judgments.

In the context of RAG ICL, the DPO metric was introduced in~\cite{contalArxiv2024} as a quantitative evaluation metric of how a retrieved document affects the LLM's probability of generating the correct response (and optionally an incorrect one) during ICL and is defined as the negative of the DPO loss~\cite{rafailovNEURIPS2023}. 

More formally, given a query \( q \), associated with a ground truth response \( r \) and optionally with an incorrect response \(\bar{r}\), and given a document \( d \) retrieved as context, the \(\mathrm{DPO}\) metric is defined as:
\begin{equation}
\label{eq:DPO}
\mathrm{DPO}(q, d) = \log \sigma \left( \log \frac{p_M\big(r \mid q, d\big)}{p_M(r \mid q)} - \log \frac{p_M\big(\bar{r} \mid q, d\big)}{p_M(\bar{r} \mid q)}\right),
\end{equation}

Where:
\begin{itemize}
    \item \( p_M(r \mid q) \) is the probability assigned by the language model \( M \) to the response \( r \) given the query \( q \) without additional context.
    \item \( p_M(r \mid q, d) \) is the probability of \( r \) when conditioned on both \( q \) and the retrieved document \( d \).
    \item \( \sigma(\cdot) \) denotes the sigmoid function.
\end{itemize}

By focusing on the ratio of probabilities, the DPO metric captures the relative increase in the likelihood of the correct response when $d$ is used as context. This aligns with the original DPO framework's goal of increasing the relative log probabilities of preferred over dispreferred responses. The DPO metric is thus used in this work to measure the quality of models used for document retrieval.

\paragraph{Computational Considerations}

Computing $\mathrm{DPO}(q, d)$ requires access to the LLM's log probabilities for both a base prompt that is only based on $q$ and an extended prompt that includes $d$ with $q$. This requires the use of LLMs with accessible probability distributions, typically open-weight models where log probabilities of the prompt can be extracted. Proprietary models without such access are unsuitable for this approach.

\section{The ICLERB Benchmark}
\label{sec:iclerb}

In this section, we introduce the In-Context Learning Embedding and Reranker Benchmark (ICLERB), a novel framework designed to evaluate retrieval methods specifically tailored for In-Context Learning (ICL) with Large Language Models (LLMs). ICLERB reframes retrieval as a ranking problem focused on enhancing LLM performance, moving beyond traditional notions of relevance or semantic similarity. In this section, we detail the benchmark framework, formalize the evaluation task and metrics, and discuss our initial implementation focused on few-shot learning scenarios, addressing the limitations of traditional benchmarks in aligning retrieval evaluation with ICL objectives.

\subsection{ICLERB Framework}

The primary objective of ICLERB is to evaluate embedding models and rerankers based on their ability to retrieve documents that enhance an LLM's performance in ICL tasks. Unlike traditional retrieval tasks that prioritize semantic relevance, ICLERB provides a standardized approach to assess the utility of retrieved documents in improving ICL outcomes by leveraging the Direct Preference Optimization (DPO) metric (Section~\ref{subsec:fine_tuning_dpo}).

Formally, consider:

\begin{itemize}
    \item A set of queries \( Q = \{ q_i \}_{i=1}^N \),
    \item A set of ground truth responses \( R = \{ r_i \}_{i=1}^N \) corresponding to the queries,
    \item A corpus of documents \( D = \{ d_j \}_{j=1}^M \),
    \item A retrieval model \( \mathcal{R}: Q \rightarrow D^M \) that, given a query \( q \) ranks the full corpus of documents \( D \).
\end{itemize}

The task is to evaluate how well the retrieval model \( \mathcal{R} \) ranks documents according to their utility in enhancing an LLM's performance, as quantified by the DPO metric.

\(\mathrm{DPO}\) possesses several desirable properties~\cite{contalArxiv2024}: It increases when the inclusion of \( d \) increases the probability of the correct answer and its magnitude is symmetric with respect to probability changes. It is also invariant to the length of the response \( r \) due to the use of probability ratios, ensuring consistency across varying response lengths and token frequencies. Finally, it is additive over independent queries, allowing for aggregate evaluation across datasets.

We note that by using DPO for a query-document pair, ICLERB focuses on the impact of retrieving a single document for ICL. While retrieving multiple documents (e.g. n-shot learning) potentially offers additional benefits, optimizing for multiple documents introduces complexities such as redundancy and diversity among the retrieved items~\cite{contalArxiv2024}, which are beyond the scope of evaluating a retrieval model.

\subsection{ICLERB Evaluation}

To benchmark the performances of retrieval models within ICLERB, we aggregate ranking metrics of cross-validation results over several datasets and several LLMs.

\paragraph{Computing Train and Test Sets.} Evaluating the performance of retrieval models within ICLERB requires constructing ground truth relevance sets based on \(\mathrm{DPO}\) (Section~\ref{subsec:fine_tuning_dpo}) for sets of test queries and documents. In order for ICLERB to correctly support approaches that require fine-tuning on \(\mathrm{DPO}\) values, we split the queries and ground truth responses dataset into a training set (80\%) and a test set (20\%). To ensure the robustness of the results, we repeat the dataset splits and experiments multiple times and report averaged metrics.
Note that most approaches evaluated in the current version of ICLERB do not make use of the training set.
We follow the framework of cold-start queries and warm-start documents, that is, all queries at test time are unseen, but the same set of documents is available during training and testing.
This emulates the typical use case where a system fine-tunes a retrieval model on a fixed set of documents, and subsequently leverages this model on unseen queries.
The ICLERB benchmark does not evaluate cold-start documents, where unseen documents are added to the corpus at test time.

Due to computational constraints, evaluating every possible query-document pair in the test set is infeasible, especially with large datasets or when utilizing high-capacity LLMs. To address this, test queries and test documents are randomly subsampled, and we then compute \(\mathrm{DPO}\) values for the quadratic number of pairs $(q, d)$ in the subsampled test sets only. After subsampling and repeated trials, our test sets require \(\mathrm{DPO}\) values for an average of $10^6$ $(q, d)$ pairs per dataset.

\paragraph{Accounting for LLM Variability.}
\(\mathrm{DPO}\) being inherently dependent on the LLM used, and to ensure robustness, ICLERB repeats the cross-validation scheme over multiple ground truth relevance sets constructed using different LLMs. Since computing \(\mathrm{DPO}\) values requires access to token probabilities, ICLERB only covers open-weight LLMs.

\paragraph{Evaluation Metrics.}
To evaluate retrieval models against the ground truth relevance sets, we compare the rankings produced by the models with the relevance induced by \(\mathrm{DPO}\) for each query. Our evaluation is based on the Normalized Discounted Cumulative Gain (nDCG) metric~\cite{jarvelinTOIS2002} to measure ranking quality. The nDCG@\(k\) metric focuses on the top \( k \) results, with higher importance placed on correctly ranking the most relevant documents at the top positions. 

Given that \( \mathrm{DPO} \) values measuring gains are negative, we rely on a modified version of the Discounted Cumulative Gain (DCG), specifically defined as follows for each query \( q \):
\begin{equation}
\label{eq:DCG}
\mathrm{DCG}@k(q) = \sum_{j=1}^{k} \frac{2^{\mathrm{DPO}(q,\mathcal{R}(q)_j)}}{\log_2(j + 1)},
\end{equation}
The nDCG is then calculated by normalizing the DCG:
\begin{equation}
\label{eq:nDCG}
\mathrm{nDCG}@k(q) = \frac{\mathrm{DCG}@k(q)}{\mathrm{IDCG}@k(q)},
\end{equation}
where \( \mathrm{IDCG}@k(q) \) is the ideal DCG value, representing the maximum possible DCG@\( k \) for a perfect ranking. We report nDCG at ranks 10 and 50, denoted as nDCG@10 and nDCG@50, respectively.

\subsection{Implementation of ICLERB for Few-Shot Learning}
\label{subsec:iclerb_for_few_shot}

In this first release of ICLERB made available through this work, we focus on evaluating retrieval methods within the context of few-shot learning for multiple choice questions (MCQ) datasets. In this setting, the corpus of documents is the set of queries and ground truth responses itself.
Note that we fix the corpus of documents after the train/test split, and the few-shot demonstrations available at both train and test time are the ground truth responses of the train set.
This specific implementation aims to assess how effectively retrievers can select demonstrations that enhance the performance of LLMs when employing ICL. 
Future versions of ICLERB will expand to include more diverse document forms and learning tasks.

\subsubsection{Few-Shot Demonstration Datasets}

MCQ demonstration datasets are constructed as a set of tuples \( \{ (q_i, a_i, A_i^-) \}_{i=1}^N \), where \( q_i \) represents a question, \( a_i \) the correct answer, and \( A_i^- \) the set of incorrect answers associated with \( q_i \). These tuples serve both as ground-truth responses and as candidate demonstrations for ICL. 

In this initial version of ICLERB, we include three datasets, selected to cover different domains and challenges and ensuring a comprehensive evaluation of retrieval methods (Table~\ref{tab:datasets_summary}):
\begin{itemize}
    \item The \textbf{TruthfulQA} dataset~\cite{Lin2021TruthfulQAMH} comprises 817 tuples aimed at assessing the capability of LLMs to generate truthful and accurate responses. This dataset focuses on questions that are prone to misconceptions or require factual correctness, evaluating the model's ability to avoid generating incorrect or misleading information.

    \item The \textbf{Emotion} dataset~\cite{saraviaEMNLP2018} consists of 20,000 tuples related to emotion recognition tasks. Each entry contains textual content expressing a particular emotion, and the task is to identify the underlying emotion from predefined categories such as \textit{joy}, \textit{sadness}, \textit{anger}, \textit{fear}, and \textit{surprise}. 

    \item The \textbf{ProductER} (Product Entity Resolution) dataset is a collection of 10,000 tuples that we manually curated, designed to showcase the practical task of product deduplication and to evaluate retrievers on a custom use case. The objective is to determine whether two product names refer to the exact same product. Each question presents a pair of product names, and the answer is categorized as \textit{yes}, \textit{no}, or \textit{maybe}, indicating whether the products are identical or not.
\end{itemize}

Questions in the ProductER dataset are formatted as:

\begin{quote}
\texttt{Product A: [Product Name A]} \\
\texttt{Product B: [Product Name B]}
\end{quote}

This dataset was designed to address the complexities inherent in product de-duplication, accounting for variations in product naming conventions, omission of important or not important details, model numbers, specifications. Using OpenAI's \texttt{o1-preview} model, we iteratively generated question-answer pairs across a range of product categories, ensuring coverage of different domains and increasing the dataset's robustness.
We aim to release the ProductER dataset on HuggingFace at \url{https://huggingface.co/crossingminds} in the near future.
 
\begin{table}[t]
\centering
\caption{Summary of datasets included in ICLERB with examples.}
\label{tab:datasets_summary}
\begin{tabular}{p{3cm} p{2cm} p{8cm}}
\toprule
\textbf{Dataset} & \textbf{Size} & \textbf{Example} \\
\midrule
TruthfulQA & 817 & \textbf{Question:} Why is Scotland part of England? \\
           &     & \textbf{Answer:} Scotland is not part of England: it's part of the United Kingdom. \\
\midrule
Emotion & 20,000 & \textbf{Text:} i have the feeling she was amused and delighted \\
         &        & \textbf{Emotion:} Joy \\
\midrule
ProductER & 10,000 & \textbf{Product A:} Bose QuietComfort 35 \\
             &        & \textbf{Product B:} Bose QC35 \\
             &        & \textbf{Answer:} Yes \\
\bottomrule
\end{tabular}
\end{table}

\subsubsection{Large Language Models Covered}

To evaluate the impact of retrieved demonstrations on ICL performance, we employ open-weight LLMs that provide access to prompt token probabilities, which are essential for computing performance metrics such as log-likelihoods and \(\mathrm{DPO}\). In this inital version of ICLERB, we use LLMs varying in size and architecture, listed below:

\begin{itemize}
    \item Mistral-7B-v0.1~\cite{jiang2023mistral7b}
    \item Meta-Llama-3.1-8B-Instruct-Turbo~\cite{dubey2024llama}
    \item Meta-Llama-3.1-70B-Instruct-Turbo~\cite{dubey2024llama}
\end{itemize}

\begin{table}[ht]
\centering
\small
\caption{ICLERB evaluation of embedding models and rerankers for ICL, sorted by nDCG@10}
\label{tab:iclerb_results}
\begin{tabular}{llccc}
\toprule
\textbf{Organization} & \textbf{Model Name} & \textbf{Model Size} & \textbf{nDCG@10} & \textbf{nDCG@50} \\
\midrule
Crossing Minds & \texttt{cm-rerank-mxbai-rlaif-v0.1} & 335M & 0.7238 & 0.7225 \\
BAAI & \texttt{bge-en-icl}~\cite{li2024makingtextembeddersfewshot, xiaoSIGIR2024} & 7.1B & 0.7191 & 0.7081 \\
NVIDIA & \texttt{NV-Embed-v2}~\cite{lee2024nv} & 7.8B & 0.7078 & 0.6998 \\
Salesforce & \texttt{SFR-Embedding-2\_R}~\cite{SFR-embedding-2} & 7.1B & 0.6925 & 0.6859 \\
DunZhang & \texttt{stella\_en\_1.5B\_v5}~\cite{zhang_stella_en_1_5B_v5_2023}  & 1.5B & 0.6885 & 0.6827 \\
NVIDIA & \texttt{NV-Retriever-v1}~\cite{moreira2024nvretrieverimprovingtextembedding} & 7.1B & 0.6878 & 0.6829 \\
Alibaba NLP & \texttt{gte-Qwen2-7B-instruct}~\cite{liCoRR2023}  & 7.6B & 0.6877 & 0.6836 \\
Cohere & \texttt{embed-english-v3.0}~\cite{cohereEmbed} & -- & 0.6876 & 0.6831 \\
Alibaba NLP & \texttt{gte-Qwen2-1.5B-instruct}~\cite{liCoRR2023}  & 1.7B & 0.6864 & 0.6825 \\
OpenAI & \texttt{text-embedding-3-large}~\cite{openai2024textembedding3small}  & -- & 0.6818 & 0.6774 \\
Linq AI Research & \texttt{Linq-Embed-Mistral}~\cite{LinqAIResearch2024} & 7.1B & 0.6793 & 0.6731 \\
Mixedbread AI & \texttt{mxbai-embed-large-v1}~\cite{emb2024mxbai,li2023angle}  & 335M & 0.6787 & 0.6782 \\
OpenAI & \texttt{text-embedding-3-small}~\cite{openai2024textembedding3small} & -- & 0.6787 & 0.6739 \\
Snowflake & \texttt{snowflake-arctic-embed-s}~\cite{merrick2024embeddingclusteringdataimprove} & 33M & 0.6714 & 0.6684 \\
Zeta Alpha AI & \texttt{Zeta-Alpha-E5-Mistral}~\cite{camara2024zetaalphae5mistral} & 7.1B & 0.6701 & 0.6672 \\
Snowflake & \texttt{snowflake-arctic-embed-l}~\cite{merrick2024embeddingclusteringdataimprove} & 334M & 0.6685 & 0.6641 \\
Voyage AI & \texttt{voyage-3-lite}~\cite{voyageaiEmbed} & -- & 0.6681 & 0.6661 \\
Cohere & \texttt{rerank-english-v3.0}~\cite{cohereRerank} & -- & 0.6679 & 0.6603 \\
SBERT & \texttt{all-MiniLM-L6-v2}~\cite{reimers-2019-sentence-bert} & 23M & 0.6672 & 0.6665 \\
SBERT & \texttt{multi-qa-distilbert-cos-v1}~\cite{reimers-2019-sentence-bert} & 66M & 0.6661 & 0.6657 \\
SBERT & \texttt{all-mpnet-base-v2}~\cite{reimers-2019-sentence-bert} & 110M & 0.6655 & 0.6649 \\
SBERT & \texttt{all-MiniLM-L12-v2}~\cite{reimers-2019-sentence-bert} & 33M & 0.6651 & 0.6646 \\
SBERT & \texttt{multi-qa-MiniLM-L6-cos-v1}~\cite{reimers-2019-sentence-bert} & 23M & 0.6650 & 0.6651 \\
Snowflake & \texttt{snowflake-arctic-embed-m-v1.5}~\cite{merrick2024embeddingclusteringdataimprove} & 109M & 0.6646 & 0.6629 \\
SBERT & \texttt{multi-qa-mpnet-base-dot-v1}~\cite{reimers-2019-sentence-bert} & 109M & 0.6644 & 0.6638 \\
Voyage AI & \texttt{rerank-2}~\cite{voyageaiRerank} & -- & 0.6386 & 0.6432 \\
\bottomrule
\end{tabular}
\end{table}

\paragraph{Implementation details.} In constructing the prompts for ICL, we follow a consistent and simple template to include the retrieved demonstration and the test question in the prompt. 
This prompt template mimics the typical structure used in few-shot learning scenarios, providing the LLM with an example to learn from. 
For each test question \( q_i \) and \( (q_{j}, a_{j}) \) is the demonstration selected by the retriever, the prompt is formulated as:

\begin{quote}
\texttt{Q:} \textit{[question $q_j$]} \\
\texttt{A:} \textit{[answer $a_j$]} \\
\texttt{Q:} \textit{[question $q_i$]} \\
\texttt{A: }
\end{quote}

To compute the log-probabilities of each answer (correct and incorrect), we then add the answer text to the prompt, get the logprobs of each token in the prompt, and extract the ones corresponding to the answer. Note that no token generation is used.

Computing logprob values was performed using the \emph{TogetherAI} API\footnote{\url{https://www.together.ai}}, which provides cloud resources to query open-weight LLMs and get token probabilities. Evaluating ICLERB from scratch involves significant computational effort. The current version involved processing approximately 1.5 billion tokens, which would take approximately 6 months at a rate of 100 tokens per second.

\subsubsection{Embedding Models and Rerankers Evaluated in ICLERB}

To assess the performance of existing retrieval models within the context of ICL, we selected state-of-the-art embedding models based on their performance in the Massive Text Embedding Benchmark (MTEB)~\cite{muennighoffEACL2023}, considered as the reference benchmark in the field. The leading models were chosen according to their ``Retrieval Average'' scores as reported in the MTEB. We also included commonly used rerankers available as a service.

The models evaluated fall into two primary categories:

\paragraph{Embedding Models.}

Embedding models (Section~\ref{subsec:related_work_text_embs_cosine_sim}) generate dense vector representations of textual data, capturing semantic information in a continuous vector space. Given a query \( q \) and a document \( d \), these models produce embeddings \( \mathbf{e}_q, \mathbf{e}_d \in \mathbb{R}^n \), where \( n \) is the dimensionality of the embedding space. The relevance score between \( q \) and \( d \) is then estimated using the cosine similarity.

\paragraph{Rerankers.}

Rerankers (Section~\ref{subsec:related_work_rankers}) are designed to refine an initial ranking of documents. Given an initial set of candidate documents (typically retrieved using an embedding model), reranking models reassess and reorder these candidates to improve relevance. These models often leverage cross-encoders or more complex architectures to capture deeper interactions between queries and documents. Typical examples include the rerankers from Cohere~\cite{cohereRerank} and Voyage AI~\cite{voyageaiRerank}.

\section{ICLERB Results}
\label{sec:iclerb_results}

This section presents the results of our evaluation of embedding models and rerankers using ICLERB and compare the overall ranking with MTEB, answering RQ1 from Section~\ref{sec:introduction}.

\subsection{Overall Results of ICLERB}

Table~\ref{tab:iclerb_results} presents the ICLERB results, sorted by nDCG@10.

The top performers on the benchmark are \texttt{cm-rerank-mxbai-rlaif-v0.1} (our proposed model, detailed in Section~\ref{sec:rlaif}), BAAI's \texttt{bge-en-icl}, and NVIDIA's \texttt{NV-Embed-v2}. These models demonstrate superior capability in retrieving contextually useful documents for ICL. Notably, the upper tier of the leaderboard is predominantly occupied by larger models, except for \texttt{cm-rerank-mxbai-rlaif-v0.1}.

Models at the lower end of the leaderboard include variants of smaller SBERT and Voyage AI's reranker. The lower performance of the rerankers from Cohere, \texttt{rerank-english-v3.0}, and Voyage AI, \texttt{rerank-2}, is particularly surprising, especially when compared to their embedding model counterparts, \texttt{embed-english-v3.0} and \texttt{voyage-3-lite} respectively, which achieve better results. This discrepancy suggests that while their embedding models are effective in capturing representations beneficial for ICL, the rerankers may be fine-tuned for other tasks like semantic similarity or search instead of ICL.

On a similar note, NVIDIA's embedding model, \texttt{NV-Embed-v2}, outperforms its retrieval model, \texttt{NV-Retriever-v1}, even though both are highly competitive and ranking at the top of the benchmark. Regarding Snowflake's models, we note that the smallest evaluated model, \texttt{snowflake-arctic-embed-s}, demonstrates better performance than its order-of-magnitude  larger counterparts, \texttt{snowflake-arctic-embed-l} and \texttt{snowflake-arctic-embed-m-v1.5}.
Both results suggests that fine-tuning higher capacity models using semantic text similarity datasets can be detrimental to ICL.

Appendix~\ref{sec:appendix_iclerb} provides additional results from ICLERB separated by datasets~(\ref{sec:appendix_ndcg_per_dataset}) and by LLMs~(\ref{sec:appendix_ndcg_per_llm}). These results show that rankings of models and rerankers significantly vary across settings. 

Typically, and looking at the performance across datasets (Section~\ref{sec:appendix_ndcg_per_dataset}), we observe that while Crossing Minds' \texttt{cm-rerank-mxbai-rlaif-v0.1} and NVIDIA's \texttt{NV-Embed-v2} maintain consistent performance across all datasets, BAAI's \texttt{bge-en-icl} shows significant variability. OpenAI's \texttt{text-embedding-3-large} and Snowflake's \texttt{snowflake-arctic-embed-m-v1.5} perform relatively better on the TruthfulQA dataset compared to their performance on the Emotion and ProductER datasets. On the other hand, \texttt{stella\_en\_1.5B\_v5} shows decreased performance on the ProductER dataset, whereas \texttt{Zeta-Alpha-E5-Mistral}, \texttt{Linq-Embed-Mistral}, and Voyage AI's \texttt{rerank-2} demonstrate improved performance on ProductER relative to their performance on other datasets.

Overall rankings across LLMs appear to be closer for Mistral-7B-v0.1 and Meta-Llama-3.1-8B-Instruct-Turbo, than for Meta-Llama-3.1-70B-Instruct-Turbo (Section~\ref{sec:appendix_ndcg_per_llm}).
Similarly, we observe again that \texttt{cm-rerank-mxbai-rlaif-v0.1} and \texttt{NV-Embed-v2} maintain consistent top performance across all LLMs, but \texttt{bge-en-icl} shows significant variability.

Future work around ICLERB will cover more datasets and LLMs, which may lead to changes in the overall rankings.

\subsection{Comparison with MTEB (RQ1)}

The Massive Text Embedding Benchmark (MTEB)~\cite{muennighoffEACL2023} is a widely used leaderboard to evaluate embedding models, in particular in the context of RAG. MTEB provides a comprehensive evaluation across tasks like classification, clustering, retrieval, and reranking, focusing on embedding quality based on these downstream tasks.

In contrast to Table~\ref{tab:iclerb_results}, Table~\ref{tab:mteb_model_summary} summarizes the performance of embedding models as reported by MTEB, sorted by the ``Average Retrieval'' score which corresponds to the standard nDCG@10. By comparing these tables, we observe some similar trends but also some discrepancies in model rankings, indicating that traditional retrieval benchmarks may not fully capture the utility of embedding models and rerankers in the context of ICL tasks.

Models such as BAAI's \texttt{bge-en-icl} and NVIDIA's \texttt{NV-Embed-v2} consistently perform well on both MTEB and ICLERB, although with slight differences in their relative rankings. Similarly, the smaller SBERT models tend to perform less favorably than most of the other models and rerankers in both benchmarks.

However, notable differences in model rankings surface when comparing MTEB and ICLERB results. Salesforce's \texttt{SFR-Embedding-2\_R}, for instance, is ranked seventh on MTEB but achieves a higher ranking on ICLERB, just below \texttt{bge-en-icl} and \texttt{NV-Embed-v2}. This suggests that \texttt{SFR-Embedding-2\_R} may be more effective for ICL tasks than traditional retrieval benchmarks indicate.

Moreover, the relatively small MixedBread AI's \texttt{mxbai-embed-large-v1} (335M parameters) performs better than \texttt{Zeta-Alpha-E5-Mistral} (7.1B parameters) on ICLERB, which is not reflected in the MTEB results. Additionally in MTEB, the three models by Snowflake are ordered accordingly to their size (33M, 109M, 334M), but on ICLERB the best model is the smallest one. This suggests that the datasets used to train these models are well calibrated for high ranks in MTEB, but not to solve ICL tasks like RAG or few-shot learning.

Overall, these observations highlight that traditional benchmarks like MTEB may not adequately reflect the nuances of retrieval models and rerankers' performances in the context of ICL. 

\section{Reinforcement Learning-to-Rank using AI Feedback}
\label{sec:rlaif}

In this section, we introduce a Reinforcement Learning-to-Rank using AI Feedback (RLRAIF) algorithm, a novel approach designed to fine-tune retrieval models for In-Context Learning (ICL). Given a set of queries and ground truth responses, and a corpus of documents, our RLRAIF algorithm explores and learns which documents maximize the improvement in LLM performance.
The reward leverages LLM-generated feedback in the form of the Direct Preference Optimization (DPO) metric (as defined in Section~\ref{subsec:fine_tuning_dpo}) to  optimize the retriever's ability to select documents that enhance an LLM's performance during ICL. 

\subsection{The Need for Reinforcement Learning in ICL Learning-to-Rank}

When fine-tuning a retriever, a primary challenge is leveraging the available data—namely, the set of queries \( Q \) associated with the set of ground truth responses \( R \), and the set of documents \( D \)—to build a suitable ranking dataset for training the retriever. Ideally, we would compute the DPO reward for all possible \( (q, d) \) pairs. However, this exhaustive computation is infeasible due to the quadratic explosion in the number of pairs, leading to unsustainable computational costs.

A naive approach involves random subsampling of \( (q, d) \) pairs. While computationally inexpensive, this method lacks focus on high-reward pairs and hard negatives, and may lead to suboptimal retriever performance given a fixed token budget. We compare the performance of this naive approach against our algorithm in Appendix Figure~\ref{fig:random_vs_rlaif}.

An improved yet still limited approach consists of using an initial unsupervised model to score all \( (q, d) \) pairs, and then employing weighted random subsampling or thresholding strategies like EPR~\cite{rubinNACL2022}. Models trained on such data inherit the limitations of the unsupervised scoring model used.

As described in Section~\ref{subsec_related_work:retrieval_models}, previous work proposes to construct training datasets by repeating the sampling strategy over a small number of iterations. For instance, the model trained at iteration \( i \), such as in UDR~\cite{liACL2023} and LLM-R~\cite{wangEACL2024}, is used to select promising \( (q, d) \) pairs to build the dataset for training the model at iteration \( i+1 \).

In the machine learning literature, such approaches of alternately querying a model to iteratively extend its own training dataset are studied under frameworks such as active learning~\cite{settles2009active}, global optimization~\cite{mockus2005bayesian}, and reinforcement learning~\cite{szepesvari2022algorithms}. In particular, efficient query acquisition algorithms must address an exploration-exploitation trade-off. Instead of acquiring rewards only for candidates that maximize predictions (pure exploitation), as done in UDR~\cite{liACL2023} and LLM-R~\cite{wangEACL2024}, it is necessary to balance exploitation with exploration to avoid getting trapped in local optima or self-reinforcement due to overconfidence.

To overcome these limitations, we propose a Reinforcement Learning-to-Rank from AI Feedback (RLRAIF) framework that frames the problem as a contextual bandit learning-to-rank problem. Like recommender systems, it is a learning-to-rank (LTR) problem since the goal is to correctly rank documents (or items, in recommender systems)~\cite{long2010active}. It is also a contextual bandit problem because the query provides the context (analogous to user context or sessions in recommender systems), and we use the model itself in the acquisition function selecting the next batch of \( (q, d) \) pairs to evaluate and obtain rewards, which serve both as our validation and training data~\cite{LiWWW2010}. We note that, compared to the contextual bandit literature, our terminology is unfortunately reversed: the ``bandit context'' corresponds to our ``query,'' and the ``bandit query,'' the documents, are placed in the ``context'' of the LLM.

Our goal is to solve a novel dual exploration-exploitation trade-off, involving both the query and document spaces. Due to the computational constraints of evaluating the reward of many \( (q, d) \) pairs, we adopt an active learning strategy to selectively sample pairs that are most informative for learning the retrieval model. This approach addresses the exploration-exploitation trade-off by prioritizing pairs balancing high expected utility with high uncertainty in reward estimation.

\subsection{RLRAIF for ICL: An Overview}

We provide below a brief overview of the main components of our RLRAIF framework that can be applied to fine-tune ranking models for ICL scenarios, without requiring an external ranking dataset. 

\paragraph{Reward Computation using LLM Feedback.}

The reward $r(q, d)$ reflects the utility of document $d$ when used as context for ICL on query $q$. We compute this reward by measuring the improvement in the LLM's output when $d$ is included as context, compared to when it is not, using the Direct Preference Optimization metric~\cite{contalArxiv2024}, \(\mathrm{DPO}\) (Section~\ref{subsec:fine_tuning_dpo}).
The objective of the reinforcement learning procedure is to minimize the simple regret of the fine-tuned ranker \( \mathcal{R} \) over all queries:
\[
\forall q \in Q,~ \max_{d \in D} r(q, d) - r(q, \mathcal{R}(q)_1)
\]

\paragraph{Pairwise Ranking Loss.}

We train the retrieval model using a pairwise ranking loss~\cite{NIPS2006_af44c4c5}, incorporating the reward signals obtained from LLM feedback. This loss function encourages the model to assign higher relevance scores to document-query pairs with higher estimated rewards.

Given a query $q$ and a pair of documents $(d^+, d^-)$ such that $r(q, d^+) > r(q, d^-)$, the pairwise ranking loss is defined as:

\begin{equation}
\mathcal{L} = \sum_{(q, d^+, d^-)} \log\left(1 + e^{- \left[ s(q, d^+) - s(q, d^-) \right]} \right),
\end{equation}

where $s(q, d)$ is the relevance score assigned by the retrieval model to the pair $(q, d)$. Minimizing this loss improves the model's ability to rank documents effectively based on their utility in enhancing LLM performance during ICL.
Compared to contrastive loss, we preserve the full ranking information and remove the need to arbitrarily convert the utility to a binary feedback.

\paragraph{Model Architecture.} Our retrieval model utilizes a cross-encoder architecture, which jointly processes the query and document texts to produce a relevance score. We start from pre-trained embeddings and fine-tune a non-linear adapter to capture complex interactions between queries and documents. Note that RLRAIF can be used to fine-tune any model architecture. The choice of fine-tuning a non-linear adapter was made due to compute resource limitations.

Let $h_q(q)$ and $h_d(d)$ represent the embeddings of the query and document obtained from two separate fine-tuned encoders. These embeddings are then combined via a cross-encoder to produce a joint embedding $h(q, d)$ that captures the interactions between the query and document. The adapter then computes the relevance score $s(q, d)$ as:

\begin{equation}
s(q, d) = \mathbf{w}^\top \sigma\left( \mathbf{W} h(q, d) + \mathbf{b} \right),
\end{equation}

where $\mathbf{W} \in \mathbb{R}^{d_1 \times d_2}$ is a weight matrix, $\mathbf{b} \in \mathbb{R}^{d_1}$ is a bias vector, $\sigma$ is a non-linear activation function (e.g., ReLU), $d_1$ is the inner embedding dimension (set at 100 in our experiments), $d_2$ is the hidden dimension of $h(q, d)$, and $\mathbf{w} \in \mathbb{R}^{d_1}$ maps the transformed features to a scalar relevance score.

\paragraph{Data Acquisition and Sampling Strategy.} To efficiently construct our training dataset within the LLM query budget, we employ an acquisition function that guides the selection of a batch of $(q, d)$ pairs based on their estimated rewards and informativeness. Our strategy includes:

\begin{itemize}
    \item \textbf{Exploitation in the Document Space:} Prioritizing documents with high expected reward, to guide the active procedure towards true optimum.
    \item \textbf{Information for Ranking Loss:} Enhancing the acquisition function with mechanisms that encourage the formation of informative document pairs per query, thus increasing the number of terms in the ranking loss.
    \item \textbf{Exploration in the Query and Document Space:} Prioritizing queries and documents that are likely to provide significant learning signals, such as those with high variance (uncertainty) in predicted rewards.
    \item \textbf{Diversity in the Batch:} Adding diversity in the batch of new pairs to evaluate avoids redundancy and maximizes mutual information.
\end{itemize}

This approach allows us to focus our computational resources on the most impactful $(q, d)$ pairs, improving the retrieval model's performance without the need to evaluate every possible pair.

\subsection{Effectiveness of RLRAIF for ICL (RQ2)}

To illustrate the effectiveness of RLRAIF and answer RQ2 (Section~\ref{sec:introduction}), we analyze its performance in the context of ICLERB. Our model \texttt{cm-rerank-mxbai-rlaif-v0.1} is fine-tuned using RLRAIF and its performance is reported in Table~\ref{tab:iclerb_results}. We initialize \texttt{cm-rerank-mxbai-rlaif-v0.1} with pre-trained embeddings from \texttt{mxbai-embed-large-v1} and fine-tune a non-linear adapter having an inner embedding dimension of 100. The adapter has approximately 150k parameters and was trained using a budget of 10k DPO values, corresponding to an estimate of 5M tokens. Our implementation leverages PyTorch and was executed on a single consumer grade GPU (NVIDIA GTX 1080 Ti).

As shown in Table~\ref{tab:iclerb_results}, \texttt{cm-rerank-mxbai-rlaif-v0.1} achieves the highest nDCG@10 and nDCG@50 scores of 0.7238 and 0.7225, respectively, outperforming all other evaluated models on ICLERB. This result indicates a significant improvement over its base embedding model, \texttt{mxbai-embed-large-v1}, which attains nDCG@10 and nDCG@50 scores of 0.6787 and 0.6782, respectively. The notable performance gain illustrates the effectiveness of RLRAIF in fine-tuning retrieval models for ICL tasks with a relatively small budget of 10k DPO values. By directly optimizing the retriever based on the LLM's performance improvements (using the DPO metric as a reward signal), RLRAIF ensures that the model's training objectives and data acquisition are intrinsically aligned with the end goal of enhancing ICL.

Comparatively, \texttt{cm-rerank-mxbai-rlaif-v0.1} surpasses larger models such as BAAI's \texttt{bge-en-icl} (7.1B parameters) and NVIDIA's \texttt{NV-Embed-v2} (7.8B parameters), which achieve nDCG@50 scores of 0.7081 and 0.6998, respectively. This suggests that fine-tuning with RLRAIF can enhance retrieval performance to a degree that compensates for smaller model sizes, emphasizing the importance of alignment with ICL objectives over sheer model capacity.

Due to resource constraints, we only fine-tuned the mixedbread AI model (\texttt{mxbai-embed-large-v1}) using RLRAIF; however, future work and iterations of ICLERB will include more performant models fine-tuned with RLRAIF.

\section{Conclusion}
\label{sec:conclusion}
In this paper, we have highlighted the limitations of using traditional retrieval methodology in the context of In-Context Learning (ICL) for Large Language Models (LLMs), arguing that retrieval for ICL should be framed as a recommendation problem rather than a search problem. We introduced a novel evaluation methodology that assesses retrieval models based on their ability to enhance LLM performance in ICL tasks, leading to the development of the In-Context Learning Embedding and Reranker Benchmark (ICLERB). Our benchmark provides a more relevant assessment of retrievers by focusing on downstream metrics that measure the utility of retrieved documents in improving LLM outputs.

Furthermore, we proposed Reinforcement Learning-to-Rank from AI Feedback (RLRAIF), a fine-tuning approach that leverages feedback from the LLM to optimize retrieval models for ICL tasks. Experimental results demonstrated that models fine-tuned with RLRAIF significantly outperform state-of-the-art retrieval models on ICLERB, even surpassing order-of-magnitude larger models optimized for traditional retrieval tasks. These findings underscore the necessity of specialized benchmarks and training strategies tailored to ICL, advocating for a paradigm shift in retrieval model development and evaluation.

Future work includes expanding ICLERB by adding more datasets, including those with more generic documents, to evaluate retrieval models across a broader range of RAG scenarios. We also plan to incorporate additional LLMs to assess the transferability of findings across different architectures and sizes. Moreover, integrating more embeddings and rerankers into the benchmark will provide a more comprehensive evaluation of retrieval strategies in the context of ICL.

\newpage
\bibliographystyle{plain}
\bibliography{references}

%%%%%%%%%%%%%%%%%%%%%%%%%%%%%%%%%%%%%%%%%%%%%%%%%%%%%%%%%%%%

\appendix

\clearpage
\section{Appendix: Supplementary Results from ICLERB}
\label{sec:appendix_iclerb}

\subsection{ICLERB Results per Dataset}
\label{sec:appendix_ndcg_per_dataset}

\subsubsection{TruthfulQA}
\begin{table}[ht]
\centering
\small
\caption{ICLERB results for TruthfulQA~\cite{Lin2021TruthfulQAMH}, sorted by nDCG@10}
\begin{tabular}{llccc}
\toprule
\textbf{Organization} & \textbf{Model Name} & \textbf{nDCG@10} & \textbf{nDCG@50} \\
\midrule
Crossing Minds & \texttt{cm-rerank-mxbai-rlaif-v0.1} & 0.6838 & 0.6821 \\
OpenAI & \texttt{text-embedding-3-large} & 0.6152 & 0.6133 \\
NVIDIA & \texttt{NV-Embed-v2} & 0.6126 & 0.6080 \\
OpenAI & \texttt{text-embedding-3-small} & 0.6122 & 0.6119 \\
Cohere & \texttt{embed-english-v3.0} & 0.6120 & 0.6103 \\
DunZhang & \texttt{stella\_en\_1.5B\_v5} & 0.6115 & 0.6089 \\
Alibaba NLP & \texttt{gte-Qwen2-1.5B-instruct} & 0.6109 & 0.6111 \\
Mixedbread AI & \texttt{mxbai-embed-large-v1} & 0.6095 & 0.6093 \\
Salesforce & \texttt{SFR-Embedding-2\_R} & 0.6092 & 0.6050 \\
Voyage AI & \texttt{voyage-3-lite} & 0.6069 & 0.6062 \\
Alibaba NLP & \texttt{gte-Qwen2-7B-instruct} & 0.6067 & 0.6089 \\
Snowflake & \texttt{snowflake-arctic-embed-m-v1.5} & 0.6063 & 0.6065 \\
SBERT & \texttt{all-mpnet-base-v2} & 0.6063 & 0.6023 \\
NVIDIA & \texttt{NV-Retriever-v1} & 0.6057 & 0.6051 \\
Snowflake & \texttt{snowflake-arctic-embed-l} & 0.6052 & 0.6013 \\
Snowflake & \texttt{snowflake-arctic-embed-s} & 0.6044 & 0.6022 \\
SBERT & \texttt{all-MiniLM-L6-v2} & 0.6029 & 0.6032 \\
Zeta Alpha AI & \texttt{Zeta-Alpha-E5-Mistral} & 0.6024 & 0.6023 \\
SBERT & \texttt{multi-qa-MiniLM-L6-cos-v1} & 0.6022 & 0.6032 \\
SBERT & \texttt{multi-qa-distilbert-cos-v1} & 0.6019 & 0.5999 \\
SBERT & \texttt{multi-qa-mpnet-base-dot-v1} & 0.6018 & 0.5995 \\
SBERT & \texttt{all-MiniLM-L12-v2} & 0.6015 & 0.6010 \\
Linq AI Research & \texttt{Linq-Embed-Mistral} & 0.6004 & 0.5959 \\
BAAI & \texttt{bge-en-icl} & 0.5917 & 0.5807 \\
Cohere & \texttt{rerank-english-v3.0} & 0.5823 & 0.5850 \\
Voyage AI & \texttt{rerank-2} & 0.5220 & 0.5463 \\
\bottomrule
\end{tabular}
\end{table}

\clearpage
\subsubsection{Emotion}
\begin{table}[ht]
\centering
\small
\caption{ICLERB results for Emotion~\cite{saraviaEMNLP2018}, sorted by nDCG@10}
\begin{tabular}{llccc}
\toprule
\textbf{Organization} & \textbf{Model Name} & \textbf{nDCG@10} & \textbf{nDCG@50} \\
\midrule
BAAI & \texttt{bge-en-icl} & 0.8251 & 0.7931 \\
NVIDIA & \texttt{NV-Embed-v2} & 0.7450 & 0.7159 \\
Crossing Minds & \texttt{cm-rerank-mxbai-rlaif-v0.1} & 0.7298 & 0.7178 \\
Salesforce & \texttt{SFR-Embedding-2\_R} & 0.7228 & 0.6988 \\
DunZhang & \texttt{stella\_en\_1.5B\_v5} & 0.7185 & 0.6916 \\
NVIDIA & \texttt{NV-Retriever-v1} & 0.7171 & 0.6925 \\
Alibaba NLP & \texttt{gte-Qwen2-7B-instruct} & 0.7129 & 0.6841 \\
Cohere & \texttt{embed-english-v3.0} & 0.7119 & 0.6886 \\
Alibaba NLP & \texttt{gte-Qwen2-1.5B-instruct} & 0.7112 & 0.6853 \\
Mixedbread AI & \texttt{mxbai-embed-large-v1} & 0.6923 & 0.6799 \\
OpenAI & \texttt{text-embedding-3-large} & 0.6897 & 0.6683 \\
OpenAI & \texttt{text-embedding-3-small} & 0.6894 & 0.6681 \\
Cohere & \texttt{rerank-english-v3.0} & 0.6878 & 0.6511 \\
Linq AI Research & \texttt{Linq-Embed-Mistral} & 0.6859 & 0.6641 \\
Snowflake & \texttt{snowflake-arctic-embed-s} & 0.6755 & 0.6539 \\
Snowflake & \texttt{snowflake-arctic-embed-l} & 0.6746 & 0.6547 \\
Voyage AI & \texttt{voyage-3-lite} & 0.6739 & 0.6585 \\
SBERT & \texttt{all-MiniLM-L6-v2} & 0.6662 & 0.6511 \\
SBERT & \texttt{multi-qa-distilbert-cos-v1} & 0.6646 & 0.6538 \\
SBERT & \texttt{multi-qa-MiniLM-L6-cos-v1} & 0.6630 & 0.6487 \\
Zeta Alpha AI & \texttt{Zeta-Alpha-E5-Mistral} & 0.6622 & 0.6445 \\
SBERT & \texttt{all-MiniLM-L12-v2} & 0.6621 & 0.6491 \\
SBERT & \texttt{multi-qa-mpnet-base-dot-v1} & 0.6618 & 0.6520 \\
Snowflake & \texttt{snowflake-arctic-embed-m-v1.5} & 0.6598 & 0.6440 \\
SBERT & \texttt{all-mpnet-base-v2} & 0.6580 & 0.6495 \\
Voyage AI & \texttt{rerank-2} & 0.6572 & 0.6346 \\
\bottomrule
\end{tabular}
\end{table}

\clearpage
\subsubsection{ProductER}
\begin{table}[ht]
\centering
\small
\caption{ICLERB results for ProductER, sorted by nDCG@10}
\begin{tabular}{llccc}
\toprule
\textbf{Organization} & \textbf{Model Name} & \textbf{nDCG@10} & \textbf{nDCG@50} \\
\midrule
NVIDIA & \texttt{NV-Embed-v2} & 0.7658 & 0.7757 \\
Crossing Minds & \texttt{cm-rerank-mxbai-rlaif-v0.1}  & 0.7578 & 0.7678 \\
Linq AI Research & \texttt{Linq-Embed-Mistral} & 0.7518 & 0.7591 \\
Zeta Alpha AI & \texttt{Zeta-Alpha-E5-Mistral} & 0.7465 & 0.7549 \\
Salesforce & \texttt{SFR-Embedding-2\_R} & 0.7456 & 0.7539 \\
Alibaba NLP & \texttt{gte-Qwen2-7B-instruct} & 0.7436 & 0.7580 \\
NVIDIA & \texttt{NV-Retriever-v1} & 0.7408 & 0.7512 \\
BAAI & \texttt{bge-en-icl} & 0.7408 & 0.7505 \\
OpenAI & \texttt{text-embedding-3-large} & 0.7406 & 0.7507 \\
Cohere & \texttt{embed-english-v3.0} & 0.7391 & 0.7503 \\
Alibaba NLP & \texttt{gte-Qwen2-1.5B-instruct} & 0.7374 & 0.7512 \\
Voyage AI & \texttt{rerank-2} & 0.7368 & 0.7486 \\
DunZhang & \texttt{stella\_en\_1.5B\_v5} & 0.7355 & 0.7478 \\
Snowflake & \texttt{snowflake-arctic-embed-s} & 0.7346 & 0.7491 \\
OpenAI & \texttt{text-embedding-3-small} & 0.7345 & 0.7419 \\
Mixedbread AI & \texttt{mxbai-embed-large-v1} & 0.7343 & 0.7455 \\
Cohere & \texttt{rerank-english-v3.0} & 0.7337 & 0.7450 \\
SBERT & \texttt{all-MiniLM-L6-v2} & 0.7326 & 0.7451 \\
SBERT & \texttt{all-mpnet-base-v2} & 0.7323 & 0.7431 \\
SBERT & \texttt{all-MiniLM-L12-v2} & 0.7317 & 0.7438 \\
SBERT & \texttt{multi-qa-distilbert-cos-v1} & 0.7317 & 0.7434 \\
SBERT & \texttt{multi-qa-MiniLM-L6-cos-v1} & 0.7299 & 0.7436 \\
SBERT & \texttt{multi-qa-mpnet-base-dot-v1} & 0.7298 & 0.7402 \\
Snowflake & \texttt{snowflake-arctic-embed-m-v1.5} & 0.7278 & 0.7385 \\
Snowflake & \texttt{snowflake-arctic-embed-l} & 0.7259 & 0.7364 \\
Voyage AI & \texttt{voyage-3-lite} & 0.7234 & 0.7333 \\
\bottomrule
\end{tabular}
\end{table}

\clearpage
\subsection{ICLERB Results per LLM}
\label{sec:appendix_ndcg_per_llm}

\subsubsection{Meta-Llama-3.1-70B-Instruct-Turbo}

\begin{table}[ht]
\centering
\small
\caption{ICLERB results for Meta-Llama-3.1-70B-Instruct-Turbo~\cite{dubey2024llama}, sorted by nDCG@10}
\begin{tabular}{llccc}
\toprule
\textbf{Organization} & \textbf{Model Name} & \textbf{nDCG@10} & \textbf{nDCG@50} \\
\midrule
Crossing Minds & \texttt{cm-rerank-mxbai-rlaif-v0.1} & 0.7205 & 0.7184 \\
NVIDIA & \texttt{NV-Embed-v2} & 0.6439 & 0.6432 \\
Voyage AI & \texttt{voyage-3-lite} & 0.6400 & 0.6409 \\
OpenAI & \texttt{text-embedding-3-large} & 0.6399 & 0.6399 \\
OpenAI & \texttt{text-embedding-3-small} & 0.6386 & 0.6387 \\
Alibaba NLP & \texttt{gte-Qwen2-7B-instruct} & 0.6379 & 0.6445 \\
Cohere & \texttt{embed-english-v3.0} & 0.6360 & 0.6371 \\
dunzhang & \texttt{stella\_en\_1.5B\_v5} & 0.6351 & 0.6403 \\
Alibaba NLP & \texttt{gte-Qwen2-1.5B-instruct} & 0.6347 & 0.6385 \\
Salesforce & \texttt{SFR-Embedding-2\_R} & 0.6321 & 0.6352 \\
mixedbread ai & \texttt{mxbai-embed-large-v1} & 0.6318 & 0.6347 \\
Zeta Alpha AI & \texttt{Zeta-Alpha-E5-Mistral} & 0.6273 & 0.6334 \\
NVIDIA & \texttt{NV-Retriever-v1} & 0.6268 & 0.6310 \\
SBERT & \texttt{all-mpnet-base-v2} & 0.6253 & 0.6250 \\
Linq AI Research & \texttt{Linq-Embed-Mistral} & 0.6252 & 0.6225 \\
Snowflake & \texttt{snowflake-arctic-embed-s} & 0.6241 & 0.6283 \\
SBERT & \texttt{multi-qa-distilbert-cos-v1} & 0.6237 & 0.6223 \\
Snowflake & \texttt{snowflake-arctic-embed-l} & 0.6234 & 0.6260 \\
Snowflake & \texttt{snowflake-arctic-embed-m-v1.5} & 0.6232 & 0.6341 \\
SBERT & \texttt{multi-qa-mpnet-base-dot-v1} & 0.6206 & 0.6251 \\
SBERT & \texttt{all-MiniLM-L6-v2} & 0.6190 & 0.6237 \\
SBERT & \texttt{multi-qa-MiniLM-L6-cos-v1} & 0.6185 & 0.6244 \\
BAAI & \texttt{bge-en-icl} & 0.6181 & 0.6052 \\
SBERT & \texttt{all-MiniLM-L12-v2} & 0.6172 & 0.6236 \\
Cohere & \texttt{rerank-english-v3.0} & 0.6011 & 0.6150 \\
Voyage AI & \texttt{rerank-2} & 0.5743 & 0.5965 \\
\bottomrule
\end{tabular}
\end{table}

\clearpage
\subsubsection{Meta-Llama-3.1-8B-Instruct-Turbo}
\begin{table}[ht]
\centering
\small
\caption{ICLERB results for Meta-Llama-3.1-8B-Instruct-Turbo~\cite{dubey2024llama}, sorted by nDCG@10}
\begin{tabular}{llccc}
\toprule
\textbf{Organization} & \textbf{Model Name} & \textbf{nDCG@10} & \textbf{nDCG@50} \\
\midrule
BAAI & \texttt{bge-en-icl} & 0.7048 & 0.6905 \\
Crossing Minds & \texttt{cm-rerank-mxbai-rlaif-v0.1} & 0.7007 & 0.7007 \\
NVIDIA & \texttt{NV-Embed-v2} & 0.6778 & 0.6700 \\
Salesforce & \texttt{SFR-Embedding-2\_R} & 0.6677 & 0.6599 \\
DunZhang & \texttt{stella\_en\_1.5B\_v5} & 0.6662 & 0.6581 \\
NVIDIA & \texttt{NV-Retriever-v1} & 0.6648 & 0.6575 \\
Cohere & \texttt{embed-english-v3.0} & 0.6643 & 0.6594 \\
Alibaba NLP & \texttt{gte-Qwen2-7B-instruct} & 0.6630 & 0.6561 \\
Alibaba NLP & \texttt{gte-Qwen2-1.5B-instruct} & 0.6625 & 0.6564 \\
OpenAI & \texttt{text-embedding-3-large} & 0.6561 & 0.6528 \\
Mixedbread AI & \texttt{mxbai-embed-large-v1} & 0.6554 & 0.6545 \\
OpenAI & \texttt{text-embedding-3-small} & 0.6553 & 0.6514 \\
Linq AI Research & \texttt{Linq-Embed-Mistral} & 0.6479 & 0.6399 \\
Voyage AI & \texttt{voyage-3-lite} & 0.6477 & 0.6463 \\
Snowflake & \texttt{snowflake-arctic-embed-s} & 0.6456 & 0.6395 \\
Cohere & \texttt{rerank-english-v3.0} & 0.6446 & 0.6316 \\
Snowflake & \texttt{snowflake-arctic-embed-l} & 0.6445 & 0.6368 \\
SBERT & \texttt{all-MiniLM-L6-v2} & 0.6417 & 0.6420 \\
SBERT & \texttt{multi-qa-MiniLM-L6-cos-v1} & 0.6407 & 0.6405 \\
SBERT & \texttt{multi-qa-distilbert-cos-v1} & 0.6401 & 0.6415 \\
SBERT & \texttt{all-mpnet-base-v2} & 0.6396 & 0.6413 \\
Snowflake & \texttt{snowflake-arctic-embed-m-v1.5} & 0.6392 & 0.6360 \\
SBERT & \texttt{all-MiniLM-L12-v2} & 0.6387 & 0.6401 \\
SBERT & \texttt{multi-qa-mpnet-base-dot-v1} & 0.6379 & 0.6404 \\
Zeta Alpha AI & \texttt{Zeta-Alpha-E5-Mistral} & 0.6376 & 0.6339 \\
Voyage AI & \texttt{rerank-2} & 0.6111 & 0.6164 \\
\bottomrule
\end{tabular}
\end{table}

\clearpage
\subsubsection{Mistral-7B-v0.1}

\begin{table}[ht]
\centering
\small
\caption{ICLERB results for Mistral-7B-v0.1~\cite{jiang2023mistral7b}, sorted by nDCG@10}
\begin{tabular}{llccc}
\toprule
\textbf{Organization} & \textbf{Model Name} & \textbf{nDCG@10} & \textbf{nDCG@50} \\
\midrule
Crossing Minds & \texttt{cm-rerank-mxbai-rlaif-v0.1} & 0.7139 & 0.7056 \\
BAAI & \texttt{bge-en-icl} & 0.6986 & 0.6793 \\
NVIDIA & \texttt{NV-Embed-v2} & 0.6840 & 0.6663 \\
Salesforce & \texttt{SFR-Embedding-2\_R} & 0.6700 & 0.6557 \\
DunZhang & \texttt{stella\_en\_1.5B\_v5} & 0.6679 & 0.6538 \\
Cohere & \texttt{embed-english-v3.0} & 0.6656 & 0.6529 \\
Alibaba NLP & \texttt{gte-Qwen2-1.5B-instruct} & 0.6653 & 0.6536 \\
NVIDIA & \texttt{NV-Retriever-v1} & 0.6643 & 0.6530 \\
Alibaba NLP & \texttt{gte-Qwen2-7B-instruct} & 0.6628 & 0.6514 \\
OpenAI & \texttt{text-embedding-3-large} & 0.6589 & 0.6459 \\
Mixedbread AI & \texttt{mxbai-embed-large-v1} & 0.6554 & 0.6487 \\
OpenAI & \texttt{text-embedding-3-small} & 0.6553 & 0.6440 \\
Linq AI Research & \texttt{Linq-Embed-Mistral} & 0.6520 & 0.6398 \\
Snowflake & \texttt{snowflake-arctic-embed-s} & 0.6469 & 0.6362 \\
Snowflake & \texttt{snowflake-arctic-embed-l} & 0.6464 & 0.6358 \\
Zeta Alpha AI & \texttt{Zeta-Alpha-E5-Mistral} & 0.6436 & 0.6343 \\
Voyage AI & \texttt{voyage-3-lite} & 0.6423 & 0.6333 \\
SBERT & \texttt{all-MiniLM-L6-v2} & 0.6421 & 0.6330 \\
Snowflake & \texttt{snowflake-arctic-embed-m-v1.5} & 0.6413 & 0.6331 \\
SBERT & \texttt{multi-qa-distilbert-cos-v1} & 0.6403 & 0.6318 \\
SBERT & \texttt{all-mpnet-base-v2} & 0.6402 & 0.6309 \\
SBERT & \texttt{all-MiniLM-L12-v2} & 0.6401 & 0.6305 \\
SBERT & \texttt{multi-qa-mpnet-base-dot-v1} & 0.6400 & 0.6299 \\
SBERT & \texttt{multi-qa-MiniLM-L6-cos-v1} & 0.6394 & 0.6320 \\
Cohere & \texttt{rerank-english-v3.0} & 0.6376 & 0.6242 \\
Voyage AI & \texttt{rerank-2} & 0.5858 & 0.5888 \\
\bottomrule
\end{tabular}
\end{table}

\clearpage
\section{Appendix: Embedding Models for Retrieval on the Massive Text Embedding Benchmark (MTEB) }

\begin{table}[ht]
\centering
\small
\caption{Embedding models as appearing in the MTEB~\cite{muennighoffEACL2023} and sorted by decreasing order of the ``Average Retrieval'' score reported there, which refers to the standard nDCG@10 metric. Table fully extracted from the MTEB benchmark on November 27th, 2024.}
\label{tab:mteb_model_summary}
\begin{tabular}{llccc}
\toprule
\textbf{Organization} & \textbf{Model Name} & \textbf{Model Size} & \textbf{Retrieval nDCG@10 from MTEB~\cite{muennighoffEACL2023}} \\
\midrule
NVIDIA & \texttt{NV-Embed-v2} & 7.8B & 0.6265 \\
BAAI & \texttt{bge-en-icl} & 7.1B & 0.6216 \\
DunZhang & \texttt{stella\_en\_1.5B\_v5} & 1.5B & 0.6101 \\
NVIDIA & \texttt{NV-Retriever-v1} & 7.1B & 0.6090 \\
Alibaba & \texttt{gte-Qwen2-7B-instruct} & 7.6B & 0.6025 \\
Linq AI Research & \texttt{Linq-Embed-Mistral} & 7.1B & 0.6019 \\
Salesforce & \texttt{SFR-Embedding-2\_R} & 7.1B & 0.6018 \\
Zeta Alpha AI & \texttt{Zeta-Alpha-E5-Mistral} & 7.1B & 0.5950 \\
Alibaba & \texttt{gte-Qwen2-1.5B-instruct} & 1.8B & 0.5829 \\
Voyage AI & \texttt{voyage-large-2-instruct} & -- & 0.5828 \\
Snowflake & \texttt{snowflake-arctic-embed-l} & 334M & 0.5598 \\
OpenAI & \texttt{text-embedding-3-large} & -- & 0.5544 \\
Snowflake & \texttt{snowflake-arctic-embed-m-v1.5} & 109M & 0.5514 \\
Cohere & \texttt{embed-english-v3.0} & -- & 0.5500 \\
Mixedbread AI & \texttt{mxbai-embed-large-v1} & 335M & 0.5439 \\
Snowflake & \texttt{snowflake-arctic-embed-s} & 33M & 0.5198 \\
OpenAI & \texttt{text-embedding-3-small} & -- & 0.5108 \\
SBERT & \texttt{all-mpnet-base-v2} & 110M & 0.4381 \\
SBERT & \texttt{all-MiniLM-L12-v2} & 33M & 0.4269 \\
SBERT & \texttt{multi-qa-MiniLM-L6-cos-v1} & 23M & 0.4117 \\
\bottomrule
\end{tabular}
\end{table}

\clearpage
\section{Appendix: RLRAIF Compared to Random Acquisition}

\begin{figure}[htbp]
    \centering
    \includegraphics[width=0.8\textwidth]{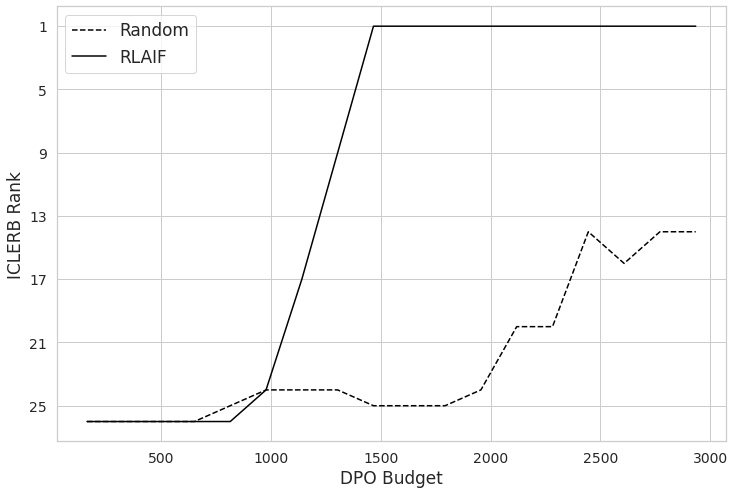}
    \caption{ICLERB rank reached with different DPO budgets, for RLRAIF and an approach randomly querying (query, document) pairs}
    \label{fig:random_vs_rlaif}
\end{figure}

\end{document}